\documentclass[10pt]{article} 

\usepackage{hyperref}
\usepackage{url}
\usepackage{graphicx}
\usepackage{xcolor}
\usepackage{color}
\usepackage{soul}
\usepackage[preprint]{tmlr}

\DeclareRobustCommand{\hlred}[1]{{\sethlcolor{pink}\hl{#1}}}

\usepackage{amsmath,amsfonts,bm}

\def\eqref#1{equation~\ref{#1}}

\def\1{\bm{1}}

\DeclareMathAlphabet{\mathsfit}{\encodingdefault}{\sfdefault}{m}{sl}
\SetMathAlphabet{\mathsfit}{bold}{\encodingdefault}{\sfdefault}{bx}{n}

\title{New Evaluation Metrics Capture Quality Degradation due to LLM Watermarking}

\author{\name Karanpartap Singh \email karanps@stanford.edu \\
      \addr Department of Electrical Engineering\\
      Stanford University
      \AND
      \name James Zou \email jamesz@stanford.edu \\
      \addr Department of Biomedical Data Science \\
      Stanford University}

\def\month{MM}  
\def\year{YYYY} 
\def\openreview{\url{https://openreview.net/forum?id=XXXX}} 

\begin{document}

\maketitle

\begin{abstract}
With the increasing use of large-language models (LLMs) like ChatGPT, watermarking has emerged as a promising approach for tracing machine-generated content. However, research on LLM watermarking often relies on simple perplexity or diversity-based measures to assess the quality of watermarked text, which can mask important limitations in watermarking. Here we introduce two new easy-to-use methods for evaluating watermarking algorithms for LLMs: 1) evaluation by LLM-judger with specific guidelines; and 2) binary classification on text embeddings to distinguish between watermarked and unwatermarked text. 
We apply these methods to characterize the effectiveness of  current watermarking techniques.  Our experiments, conducted across various datasets, reveal that current watermarking methods are detectable by even simple classifiers, challenging the notion of watermarking subtlety. We also found, through the LLM judger, that watermarking impacts text quality, especially in degrading the coherence and depth of the response. Our findings underscore the trade-off between watermark robustness and text quality and highlight the importance of having more informative metrics to assess watermarking quality. 
\end{abstract}

\section{Introduction}

The advancement of Large Language Models (LLMs) like GPT-4 and Llama-2 has heralded a new era in natural language processing, offering unprecedented capabilities in generating human-like text \citep{OpenAI2023GPT-4Card, Touvron2023LlamaModels, Chowdhery2022PaLM:Pathways}. However, this advancement also brings forth a unique challenge: ensuring the integrity and traceability of machine-generated content \citep{Clark2021AllText, Mora-Cantallops2021TraceabilityTools}. These concerns have led to the development of many watermarking techniques for LLMs, aimed at embedding identifiable markers into generated text without compromising its quality or readability \citep{Kirchenbauer2023AModels, Kirchenbauer2023OnModels, Takezawa2023NecessaryModels, Zhao2023ProvableText, Christ2023UndetectableModels, Yoo2023RobustFeatures, Zhao2023ProtectingWatermarking}.

Watermarking in the context of LLMs is a relatively new and rapidly evolving field. The primary objective is to embed a non-obtrusive, detectable marker within the text generated by these models, enabling the identification of the source model and potentially deterring misuse such as plagiarism or misinformation. Recent advances have introduced sophisticated techniques aimed at embedding watermarks seamlessly into the language model's output through black-box approaches, ensuring minimal impact on the text's quality without access to the original model's weights. However, the effectiveness and subtlety of these methods remain under scrutiny \citep{Tang2023TheTexts}. Key concerns include the detectability of these watermarks by third parties, their potential to degrade text quality, and the challenge of maintaining the watermark's integrity across different contexts and content types. Prior work has used metrics such as perplexity \citep{Kirchenbauer2023AModels}, n-gram log diversity \citep{Kirchenbauer2023OnModels}, or BLEU scores for machine-translation tests \citep{Takezawa2023NecessaryModels} to evaluate the quality impact of watermarking. However, these metrics fail to capture factors such as the semantic coherence or contextual relevance of the generated text.

In this work, we propose two new benchmarks to assess watermarking algorithms for large language models. We focus our work on determining whether an independent classifier can distinguish a generation as being watermarked without prior knowledge about the watermarking algorithm or associated secret keys. Through experimentation across various algorithms and datasets, we demonstrate that current watermarking methods can indeed be detected by such classifiers. This finding challenges the prevailing notion of watermark subtlety and calls for a reevaluation of current techniques. By exploring the robustness and quality impact of different watermarking algorithms, we aim to advance the field towards developing more effective and less intrusive watermarking solutions for LLMs. 

\begin{figure*}[t]
\includegraphics[width=\textwidth]{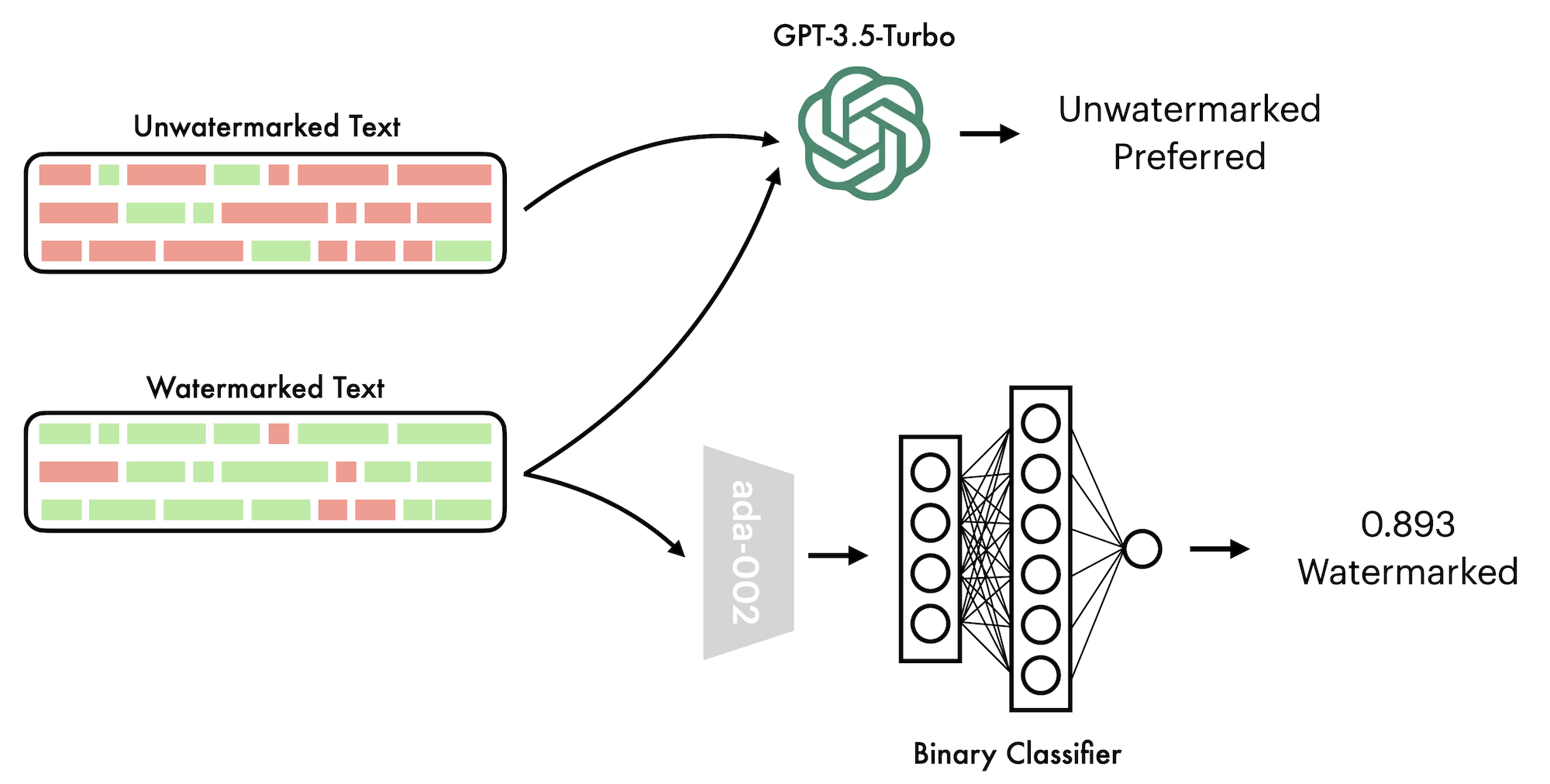}
\caption{Can watermarked outputs from large language models be distinguished with a black-box approach? We answer this question through two new methods for evaluating LLM watermarks, showing that independent classifiers and judgers with no prior knowledge of watermarking algorithms prefer or can effectively classify watermarked outputs.}
\end{figure*}

\section{Methods}

\subsection{Evaluation Mechanisms}

We present two evaluation mechanisms to assess the robustness and quality of the outputs generated by the watermarked models:

\begin{enumerate}
    \item \textbf{Automated Judgement with GPT}: We employed GPT-3.5-Turbo with a tailored prompt to act as an impartial judge, inspired by \citet{Zheng2023JudgingArena}, and rank generated outputs (watermarked and unwatermarked) on a 1-5 Likert scale for the following factors, selecting by prompting GPT-4 for an appropriate set of criteria for assessing language model generations: relevance to the prompt, depth of detail, clarity of writing, coherence and logical flow, originality and insight, use of specific examples, and accuracy of information. We chose GPT-3.5-Turbo for our primary analyses because of its balance of capabilities and cost/accessibility. However, we also conducted a comparison study between the detailed prompt with Llama-2-7B-chat, GPT-3.5-Turbo, and GPT-4, as well as a simpler prompt without the categorical ranking system with GPT-3.5-Turbo. \newline
    
    The prompt instructed the judger to provide specific examples and reasoning for its scoring, as well as a final verdict for which output it preferred overall. Lastly, to account for any positional biases inherent to GPT \citep{Wang2023LargeEvaluators}, we randomized the order of the outputs presented to the judger. The full prompts and representative responses are included in Appendix \ref{sec:judger_details}.
    
    \item \textbf{Binary Classifier}: Based on text embeddings obtained using OpenAI's \texttt{text-embedding-ada-002} model from the two outputs (watermarked and unwatermarked), a simple multi-layer perceptron (MLP)-based binary classifier, consisting of 961 neurons in 4 layers, was trained to classify a given text as either unwatermarked or watermarked. A hyperparameter search was performed for each experiment and dataset to ensure the best performance. More details regarding the network and its training are presented in Appendix \ref{sec:classifier_details}. Simple logistic regression on the same embeddings was also tested as a classifier. When applicable, k-fold cross-validation was performed with random shuffling and 5 folds.
\end{enumerate}

\subsection{Datasets}

Three datasets were tested in this study. For all datasets, a section of text up to 50 words long was spliced from each sample, after which the 7 billion parameter variant of the Llama-2 model was tasked with completing the output \citep{Touvron2023LlamaModels}, both with and without a watermarking layer applied.  

\begin{enumerate}
    \item \textbf{LongForm, Validation Set}: we used the Wikipedia subset of the LongForm dataset's validation set, consisting of 251 human-written documents on various topics \citep{Koksal2023LongForm:Extraction}.
    \item \textbf{C4-RealNewsLike, Validation Set}: A subset of the Common Crawl web crawl corpus, the RealNewsLike dataset contains text extracted from online news articles \citep{Raffel2019ExploringTransformer}. We used 500 samples from this dataset.
    \item \textbf{Scientific Papers, Test Set}: A collection of long, structured documents from the arXiv and PubMed open access article repositories \citep{Cohan2018ADocuments}. We used the abstracts from 252 samples.
\end{enumerate}

\subsection{Watermarking Techniques}
We examined two distinct watermarking techniques, the Soft-Watermark \citep{Kirchenbauer2023AModels} and Robust Distortion-Free Watermark \citep{Kuditipudi2023RobustModels}, in this study, though our evaluation metrics could be applied to any watermark using a black-box approach, regardless of its complexity or other attributes. In selecting these techniques, we prioritized methods that were readily reproducible with published codebases. We study the soft-watermark for our primary analysis because of its computational efficiency, popularity, and the similarity of many subsequent methods to it. Additionally, we also study a very recent distortion-free watermarking technique to illustrate how our metrics can reveal limitations in watermarking methods that explicitly aim to maintain the original quality in the generated text. 

\begin{enumerate}
\item \textbf{Soft-Watermarking}: Soft-watermarking, involves the pre-selection of pseudo-random "green" tokens before word generation, promoting their use during sampling \citep{Kirchenbauer2023AModels}. This watermark can be detected through a statistical test with knowledge of the secret key and token hashing function used to select the green list. This technique was optimized for improved reliability in Kirchenbauer et al., and we use the updated SelfHash scheme in all of our experiments \citep{Kirchenbauer2023OnModels}. The watermarking parameters remain fixed at $(\gamma, \delta) = (0.25, 4.0)$ unless otherwise stated. These parameters were chosen using recommendations from \citet{Kirchenbauer2023AModels}, along with testing on Llama-2-7B to ensure that watermarks with sufficiently high z-scores (z=4) were implanted.

\item \textbf{Robust, Distortion-Free Watermarking}: This watermarking technique, involves using inverse transform sampling or exponential minimum sampling to embed a watermark into the output of a language model \citep{Kuditipudi2023RobustModels}. This is achieved by mapping a sequence of random numbers from a watermark key to the probabilities assigned by the language model to each possible next token in the text. The watermarked text thus encodes information about its source that can be detected by aligning the sequence of tokens with the known sequence of random numbers from the watermark key. This method is meant to be distortion-free by leveraging the inherent randomness in the language model's text generation process, ensuring that the introduction of a watermark does not change the distribution of the generated text. We use the most robust default (EXP-edit) variant of the watermark provided in the open-source implementation, with the default parameters ($n=256$ for length of the watermark sequence, and $\text{key}=42$ for the secret key). We note that, on the same hardware (1x NVIDIA Tesla GPU), this watermark took between 15-20x the computation time for each generation as compared to the original model or soft-watermark.

\end{enumerate}

\section{Experiments}

\begin{figure*}[h]
\includegraphics[width=\textwidth]{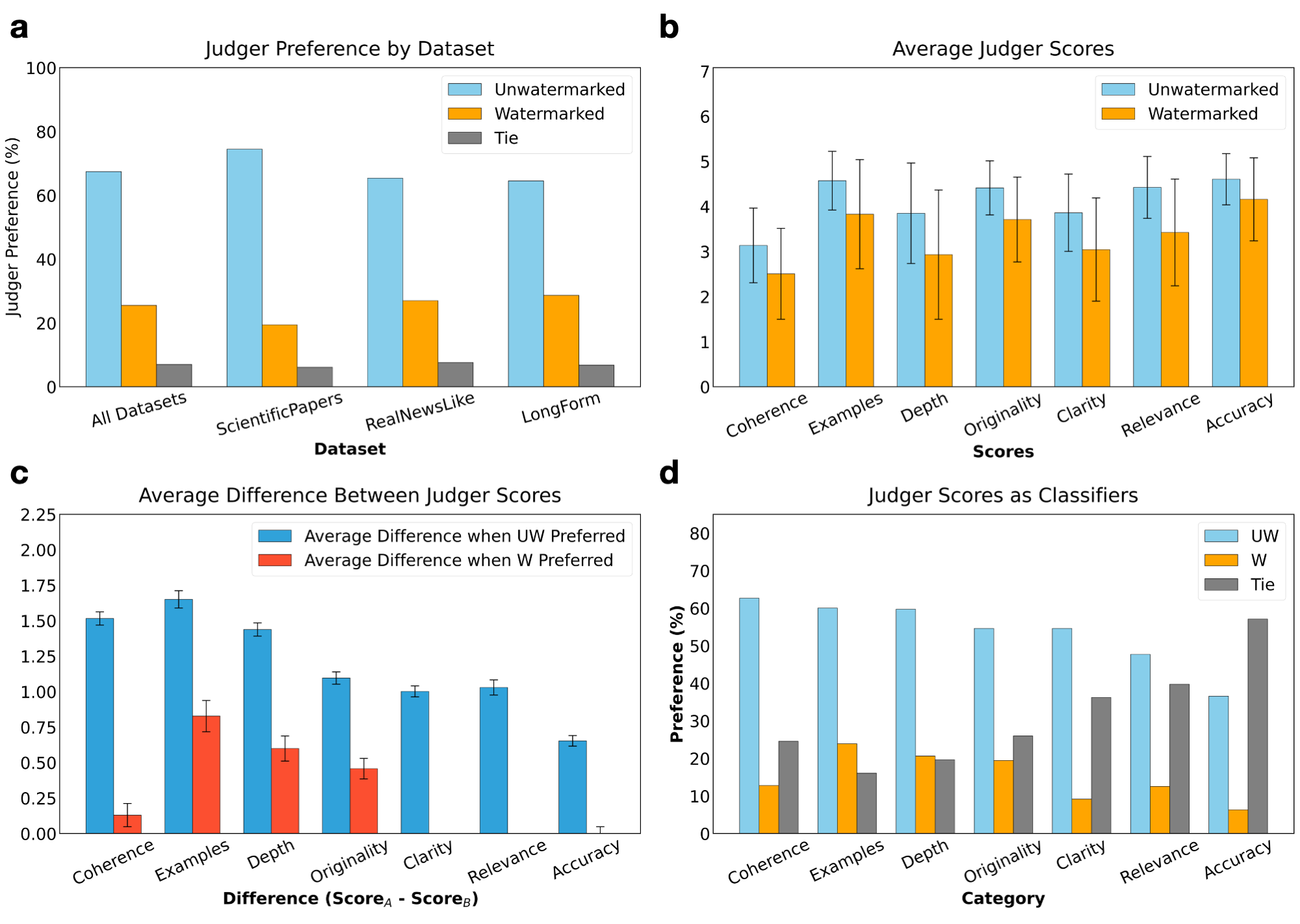}
\label{judger_results}
\caption{a) GPT-judger preference for the soft-watermark (given as the percentage of samples preferred for each class) for each dataset, separated into samples where unwatermarked outputs were preferred, watermarked samples were preferred, or neither was preferred (tie). b) Average scores for each of the 7 evaluation categories provided to the judger for all unwatermarked and watermarked samples (N = 1003). Error bars indicate the standard deviation in the scores. c) Average score differences for each evaluation category when either unwatermarked or watermarked samples were preferred. d) Judger preference when using the scores in each category as a classifier. Categories are ordered by the highest preference for unwatermarked samples as compared to watermarked samples or ties.}
\end{figure*}

\subsection{Detectability of Watermarked Text}

Upon evaluation by the GPT-judger, across all samples, the unwatermarked outputs were preferred approximately 67.4\% of the time, with the watermarked outputs being preferred 25.5\% of the time. The remaining samples were declared a tie by the judger. This trend held for each individual dataset as well, with the unwatermarked text completions being declared better responses for between 64.5\% and 74.5\% of the samples (Figure 2a). These results indicate the independent judger's ability to distinguish between unwatermarked and soft-watermarked outputs, pointing to quality degradation as a result of the watermark, or other effects of watermarking on the LLM's outputs. To extract more insight from the results of the GPT-judger, we also examine the scores given by the judger to each output. 

\subsection{Judger Reasoning}

Across the judging criteria provided in the prompt, the GPT-judger, on average, gave a higher score to the unwatermarked text completions (Figure 2b). However, the range of these scores was relatively large. Seeking to determine the reasons why certain unwatermarked outputs were preferred to their watermarked counterparts, we looked at both the average difference in the scores of the two when either one was preferred, as well as used the scores in each category as a classifier. In over 50\% of samples, the accuracy of both text completions was determined to be equal by the judger, indicating that the watermark does not affect this attribute of the resulting text (to the best evaluation capabilities of the judger) (Figure 2c). However, a large difference was seen in the coherence scores given to the texts, with both the average differences in the scores and judger preference for this category being significantly higher for the unwatermarked samples (Figure 2c-d). Similar results can be seen for the "use of examples" and "depth" categories, indicating that soft-watermarking tends to negatively affect these attributes of the generated text.

\begin{figure}[t]
\includegraphics[width=\textwidth]{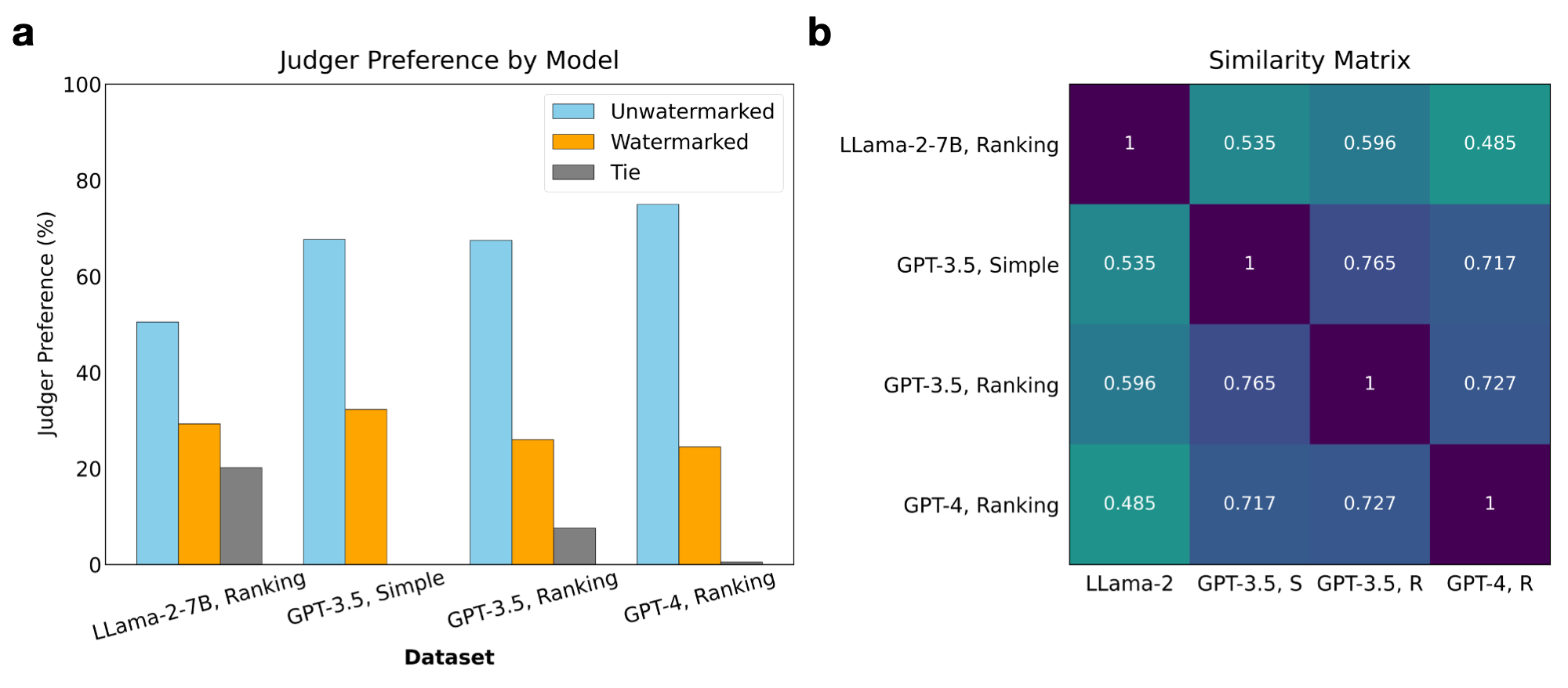}
\label{judger_comparison}
\caption{a) Judger preference when evaluating LLama-2-7B, GPT-3.5-Turbo, and GPT-4, with categorical ranking-based and simple prompts. All results are for a subset of N=200 samples from the RealNewsLike dataset. b) Similarities, given as the percentage of classifications that were the same, between all judgers. The models on the x-axis are abbreviated for brevity but follow the same sequence as those on the y-axis.}
\end{figure}

\subsection{Judger Comparison}

To assess the impact of judger model strength on its performance, as well as to determine the impact of the category-based ranking system in the prompt, we evaluated four different setups: Llama-2-7B-chat with the ranking-based prompt, GPT-3.5-Turbo with a simpler prompt without categorical ranking, GPT-3.5-Turbo with the ranking prompt, and lastly, GPT-4 with the ranking prompt. Both prompts are included in Appendix \ref{sec:judger_details}. 

The results, shown in Figure 3, indicate that even a small model, Llama-2-7B, shows a preference for the unwatermarked model versus the watermarked model. It is worth noting that Llama-2 was significantly worse at adhering to the result format required by the prompt, and required repeated sampling to obtain parseable results. Upgrading to GPT-3.5-Turbo, the simpler prompt, which only required the judger to output a binary verdict for its preferred completion with no reasoning, delivered nearly identical results to the ranking-based prompt. However, the addition of the categorical ranking allowed us to extract further insights from the judger and determine which aspects of the generated text suffered the most from watermarking. GPT-4 showed an even higher preference of 75\% for the unwatermarked completions, however at a substantial increase in cost. These results indicate that larger models are more discerning of the differences between unwatermarked and watermarked outputs, and may extend to newer and more capable models such as GPT-4-Turbo. 

All of the GPT-based judgers agreed with each other over 71\% of the time, with the highest agreement of 76.5\% being between the GPT-3.5-Turbo results for the simple and categorical ranking prompts. The Llama-2-based judger agreed with the GPT-based judgers for roughly half of the samples.

\newpage \subsection{Binary Classifier Performance}

\begin{figure}[h]
\includegraphics[width=\textwidth]{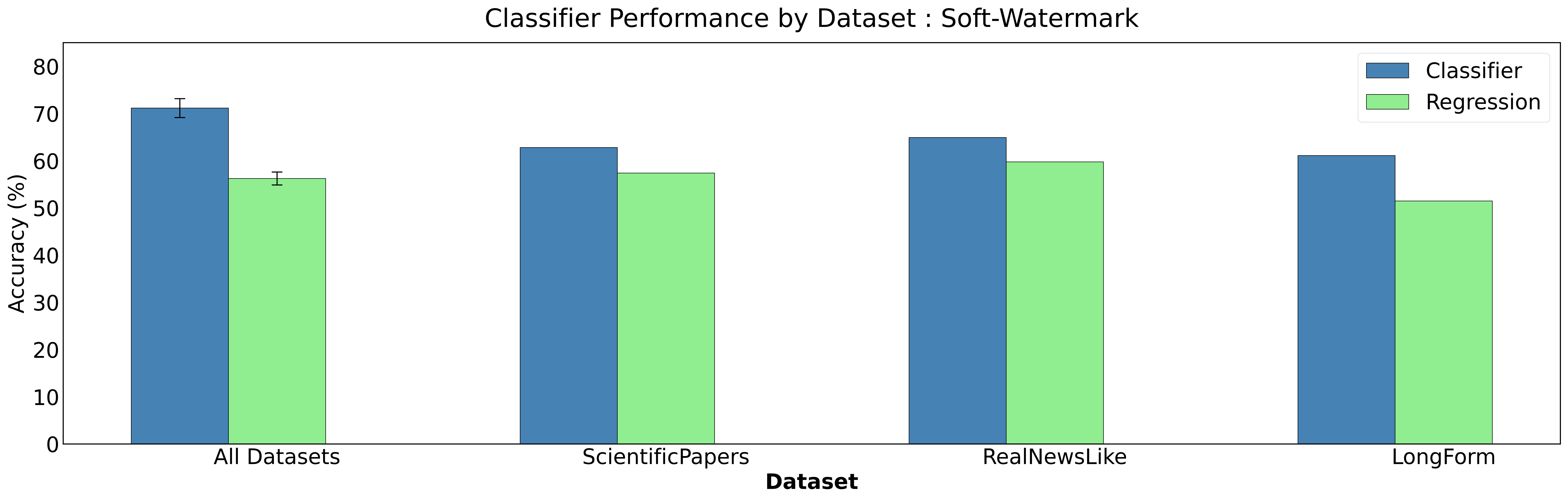}
\label{classifier_results}
\caption{Binary classifier (neural-network based and regression) performance on each dataset for the soft-watermark. When evaluating all of the datasets together, k-fold cross-validation was used with 5-folds. For the three individual datasets, each algorithm was trained on the indicated dataset, and tested on the other two datasets for generalizability.}
\end{figure}

To further test whether watermarking discernibly altered the generated text, we employed two classification methods: a 4-layer neural network (NN) and simple logistic regression. These classifiers were trained to distinguish between watermarked and unwatermarked texts using \texttt{text-embedding-ada-002} embeddings obtained from both texts.

The neural network classifier showed promising results, yielding an average accuracy of 71\% when trained on all of the datasets, along with an AUC of 0.75, false-unwatermarked rate of 30.8\%, and false-watermarked rate of 30.4\%. This high accuracy was consistent, even when the model was trained on one dataset and tested on the other two, highlighting the network's ability to generalize and recognize watermarking patterns without explicit knowledge of the watermarking techniques or access to the secret key (Figure 4).

Logistic regression, a simpler classification method, was also employed as an ablation study to determine if watermark detection was feasible with more basic techniques. This classifier achieved an accuracy just above random guessing, at approximately 56\%, across various datasets using k-fold cross-validation with 5 folds. Notably, its highest accuracy, 60\%, was observed when trained on the RNL (RealNewsLike) dataset and tested on the others, likely due to the larger sample size of 500 in the RNL dataset. Despite logistic regression's relatively lower performance compared to the MLP-based classifier, these results still indicate the presence of a detectable watermarking signal in the texts (Figure 4).

\subsection{Detectability vs. Watermarking Parameters}

\begin{figure*}[h]
\includegraphics[width=\textwidth]{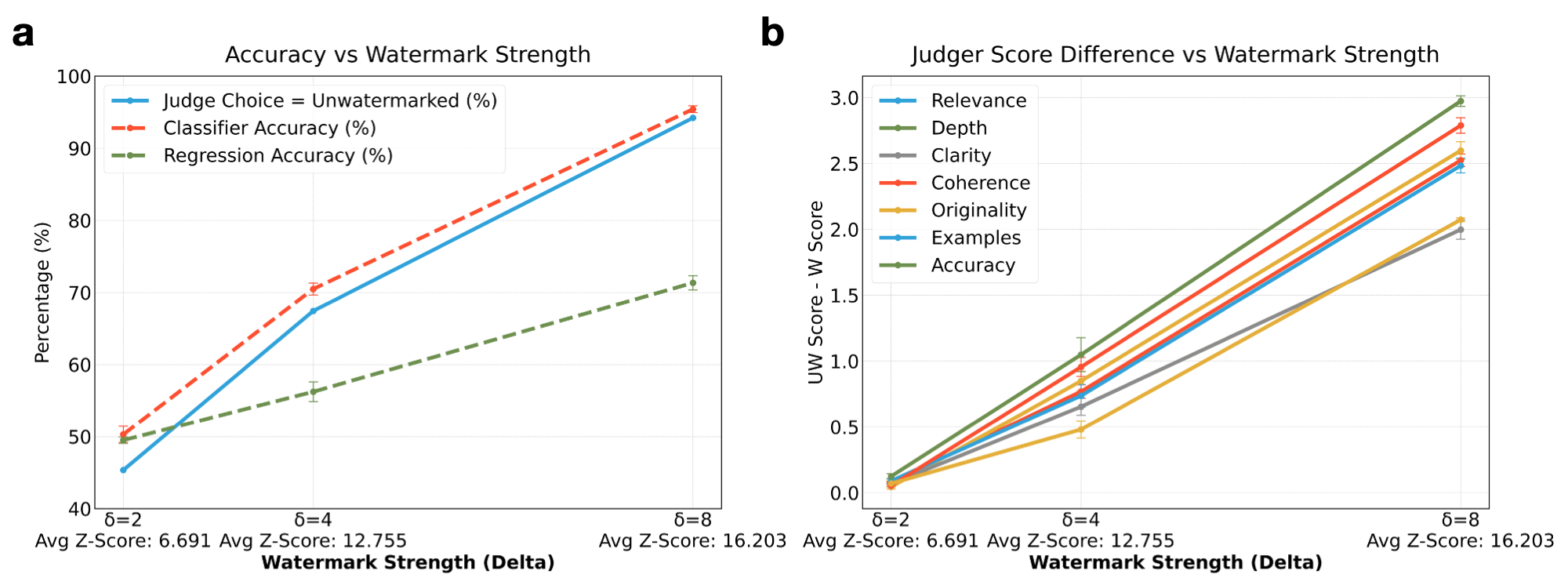}
\label{accuracy_vs_strength}
\caption{a) Judger preference, classifier accuracy, and regression accuracy for various soft-watermark strengths ($\delta=(2, 4, 8)$), evaluated across all of the datasets with k-fold cross-validation (5 folds). Larger $\delta$ corresponds to stronger watermarking. b) Average judger score differences for each category when varying the watermark strength. (All) error-bars represent the standard error in the measurement.}
\end{figure*}

Watermarks for large language models present an inherent trade-off between robustness, computational cost, and text quality \citep{Kirchenbauer2023AModels, Kirchenbauer2023OnModels}. We therefore evaluated the impact of the watermark strength on the accuracy of our techniques. For both of the datasets we tested, the judger's preference for the unwatermarked output and classifier's accuracy in predicting whether an output was watermarked increased as the watermark was made stronger (Figure 5a). At the lowest tested delta parameter of $\delta=2$, the watermark was still robust enough to be detected, but was not discernible to a high degree by the judger or classifier. Additionally, as the watermarking strength was increased, the difference between the judger's scores for the unwatermarked and watermarked outputs also increased in all categories, with the largest difference being seen for the depth of the outputs (Figure 5b). 

\begin{figure}[t]
\includegraphics[width=\textwidth]{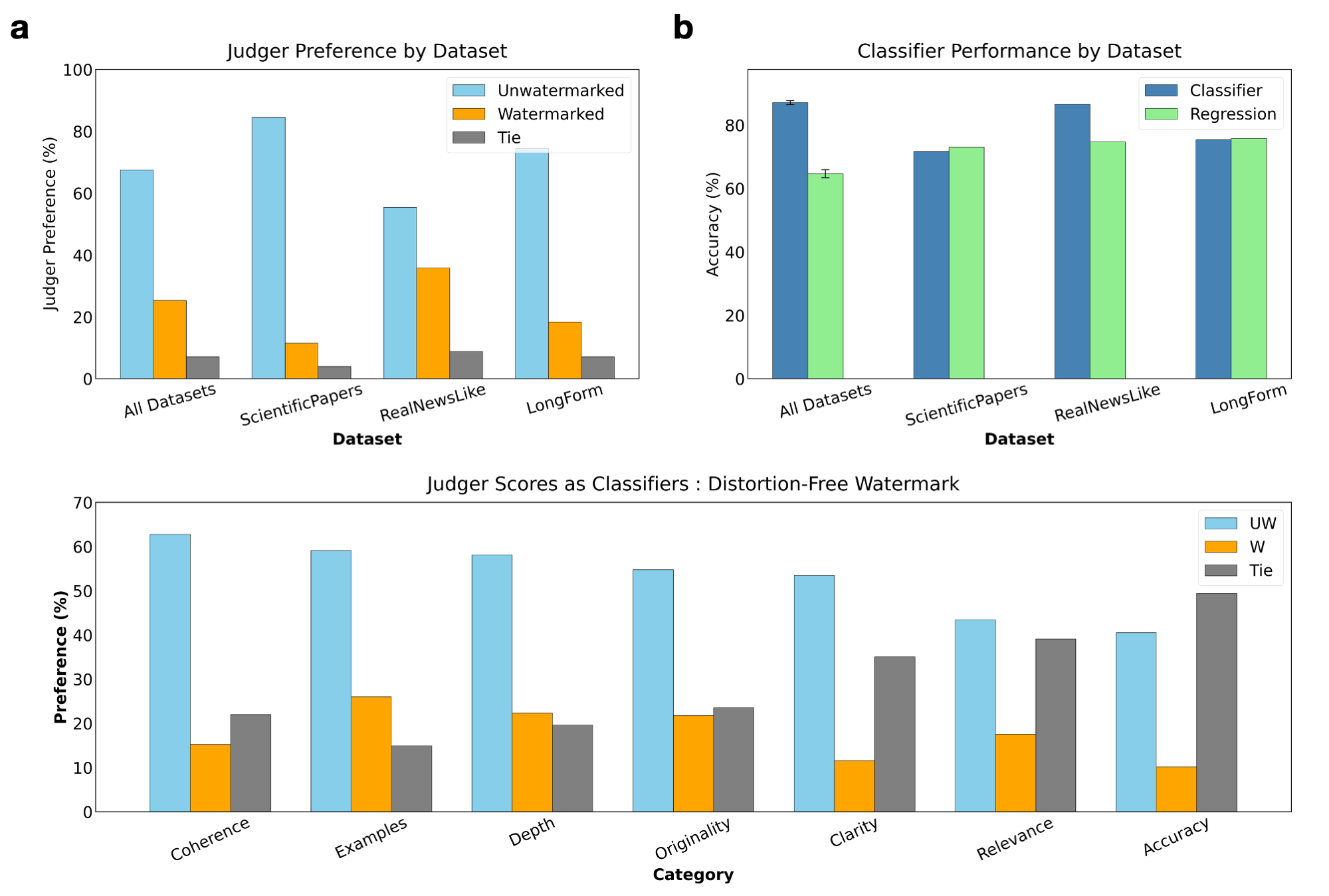}
\label{distortion_free}
\caption{a) GPT-judger preference for the distortion-free watermark (given as the percentage of samples preferred for each class) for each dataset, separated into samples where unwatermarked outputs were preferred, watermarked samples were preferred, or neither was preferred (tie). b) Binary classifier (neural-network based and regression) performance on each dataset for the distortion-free watermark. k-fold cross-validation was used with 5-folds when evaluating all of the datasets. For the three individual datasets, each algorithm was trained on the indicated dataset, and tested on the other two. c) Judger preference when using the scores in each category as a classifier.}
\end{figure}

\subsection{Evaluating Robust Distortion-Free Watermark}

Lastly, we evaluated both techniques on another watermarking algorithm, the Robust Distortion-Free Watermark proposed by Kuditipudi et. al \citep{Kuditipudi2023RobustModels}, to test their generalization. The judger showed a very similar preference of 67.5\% for the unwatermarked samples and 25.3\% for watermarked samples, with the remaining 7.2\% of samples being declared ties (Figure 6a). On the ScientificPapers dataset, the judger preferred 84\% of the unwatermarked samples, indicating an even larger degradation in quality for this class of academic text generation. Once again, the judger's decisions were most influenced by the lack in coherence, depth, and usage of examples in the watermarked text (Figure 6c).

Meanwhile, the MLP-based classifier achieved above 80\% accuracy and 0.926 AUC in discerning watermarked samples when evaluated on all of the datasets, with linear regression also displaying a detection accuracy near 75\% when trained on each individual dataset (Figure 6b). Collectively, these results indicate that the robust, distortion-free watermark also perceptibly affects generation quality.

\section{Discussion}

\paragraph{Related Works} In the evolving landscape of large language model applications, watermarking has emerged as a crucial technique for tracing model outputs. Most current LLM watermarking approaches involve subtly biasing the model's logits using pseudorandom distributions. These techniques vary, ranging from simple binary partitioning with 'green' and 'red' lists to more sophisticated methods. In parallel, techniques like the LLM-judger have emerged as valuable tools for comparing and benchmarking LLMs \citep{Zheng2023JudgingArena}, or generating annotations for instruction tuning \citep{Wang2023PandaLM:Optimization, Peng2023InstructionGPT-4, Zhou2023LIMA:Alignment}. 

In this study, we introduced two new techniques for evaluating watermarks for large language models. Fundamentally, watermarking for LLMs should remain invisible to both automated systems and human evaluators. Our study reveals that, contrary to this ideal, current watermarking techniques, including soft-watermarking and the EXP watermark, introduce detectable patterns or anomalies into the generated text. The ability of independent classifiers to detect watermarked content in LLMs without prior knowledge of the specific watermarking algorithm or secret keys is a notable finding. The effectiveness of simple classifiers like logistic regression and multi-layer perceptrons, achieving over 70\% and up to 86.5\% accuracy in identifying watermarked content, further underscores this point. This finding not only questions the non-detectability of these watermarks but also suggests that in practice, even methods designed to be distortion-free still suffer from degradations.

The GPT-judger's scoring further adds a dimension to this understanding by highlighting the specific areas of text quality that are impacted by current watermarking methods. Watermarking, as we observed, tends to degrade text attributes like coherence and depth, whereas internal accuracy remains generally consistent. This degradation is crucial as it can compromise the utility and acceptability of watermarked texts in settings where high-quality outputs are paramount. Our results also shed light on the trade-off between watermark strength and text quality. As watermark strength increases, so does its detectability, indicating more pronounced quality degradation. With a particularly robust watermark, we see detection accuracies over 90\%, all without any priors regarding the watermarking algorithm and with relatively small sample sizes used for training.

The ability to detect watermarks in texts generated using various watermarking methods suggests that detectability might be a crucial characteristic to consider when developing these methods. Looking ahead, our work opens new avenues for research. Future studies could employ more sophisticated classifiers, delve deeper into the nuances of how watermarking alters text generation, and ultimately use these insights to develop new watermarking techniques that do not alter the generated text perceptibly from the original model. 

In conclusion, our findings provide an overview of the current state of watermarking in LLMs. The challenges in achieving undetectability and maintaining text quality are more pronounced than previously understood. As LLMs become increasingly prevalent, developing watermarking techniques that are both robust and subtle is crucial.





\bibliography{main}
\bibliographystyle{tmlr}

\appendix
\newpage \section{Appendix 1: Judger}

\label{sec:judger_details}

\textbf{Example Response Set:}

\begin{table}[h]
\caption{Example response set for the soft and distortion-free watermarks. Incoherent sections of the watermarked outputs that likely contribute to the judger's decision are highlighted. For longer such sections, only the first sentence is highlighted for visibility. DF-Watermark refers to the distortion-free watermark.}
\label{example_responses}
\begin{center}
\begin{tabular}{p{0.2\textwidth}p{0.2\textwidth}p{0.2\textwidth}p{0.2\textwidth}}
\multicolumn{1}{c}{\bf PROMPT}  & \multicolumn{1}{c}{\bf UNWATERMARKED} & \multicolumn{1}{c}{\bf SOFT-WATERMARK} & \multicolumn{1}{c}{\bf DF-WATERMARK}
\\ \hline \\
A little over 5 acres of land in a prime area of Western North Carolina. Upon arrival to the home site, you'll pass gorgeous mountain views, a babbling creek and pasture land. This property has a storage shed already in place and the land is ready for your use which

& makes it easy to start building your dream home. The property is located in a quiet and peaceful area with a short distance to the Blue Ridge Parkway and all the outdoor activities it has to offer. The nearby town of Spruce Pine offers shopping, dining and entertainment options. Don't miss out on this amazing opportunity to own a little piece of heaven. \newline

Directions: From the intersection of Hwy 221 and Hwy 70 in Spruce Pine, take Hwy 70 East for approximately 4 miles. Property will be on the right just past the Blue Ridge Parkway Overlook. Look for the sign.

& could be a homestead, farm, or a place to park your RV. Close to the Blue Ridge parkway, Boone and Watauga river making this a prime location for outdoor enthusiasts or just a great place to unwind. Property is being sold as is with no disclosures or warranties. \hlred{Seller would like a 30 day inspection period}, and a cash deal would be the most likely outcome, however would discuss with the right offer. \hlred{Agent is related to the seller.} This is a great chance to live the country life close to everything. 

& includes double rural zoning. With a little creativity and some elbow grease this property can be turned into your very own piece of paradise. Sunset views are nothing short of breathtaking. Location is just minutes from the nearby cities of Marshall, Bryson City and Hayesville. This property is priced to sell and won't last long. Call today and schedule a viewing! \hlred{When we first meet Scott Morrison, he is sitting in his office.} He is wearing a sky blue shirt with a hint of a smile, looking worn and exhausted. He has been awake for days... \newline

Scott Morrison is the 30th Prime Minister of Australia, serving since August 2018. Since his appointment, he has been at the forefront of national issues, from climate change to the economy..., and dealing with the fallout of the Kavenaugh affair.\\
\end{tabular}
\end{center}
\end{table}

\newpage \textbf{Judger Prompt and Example Output:}

\begin{table}[h]
\caption{Judger prompt and completion from GPT-3.5-Turbo for the example response set given in Table \ref{example_responses}. In this example, the randomization resulted in LLM A being the watermarked completion, and LLM B being the unwatermarked completion.}
\label{judger_example}
\begin{center}
\begin{tabular}{p{0.5\textwidth}p{0.5\textwidth}}
\multicolumn{1}{c}{\bf PROMPT}  & \multicolumn{1}{c}{\bf JUDGER RESPONSE}
\\ \hline \\
Please act as an impartial judge and evaluate the quality of the text completions provided by two large language models to the prompt displayed below. Assess each response according to the criteria outlined, using a 1-5 Likert scale. After scoring each criterion, provide a summary for each response, including examples that influenced your scoring. Additionally, ensure that the order in which the responses are presented does not affect your decision. \newline

\textbf{Criteria:}
\begin{enumerate}
    \item Relevance to the prompt
    \item Depth of detail
    \item Clarity of writing
    \item Coherence and logical flow
    \item Originality and insight
    \item Use of specific examples
    \item Accuracy of information
\end{enumerate}

Start with a brief statement about which response you think is superior. Then, for each response and criterion, provide a score, followed by a brief justification for that score. At the very end of your response, declare your verdict by choosing one of the choices below, strictly following the given format: \newline
    
"[[A]]: [list of scores]" if assistant A is better, \newline
"[[B]]: [list of scores]" if assistant B is better, or \newline
"[[C]]" for a tie. \newline

[Prompt] \newline

[LLM A’s Answer] \newline

[LLM B’s Answer] \newline

& In my assessment, the superior response is from LLM B. \newline

1. Relevance to the prompt: \newline
   - LLM A: 3 \newline
   - LLM B: 5 \newline

   LLM B's response directly addresses the prompt by emphasizing the opportunity to build a dream home and providing directions to the property. In contrast, LLM A's response focuses more on general information about the location and potential uses of the land, which is somewhat relevant but not as specific. \newline

2. Depth of detail:  \newline
   - LLM A: 4  \newline
   - LLM B: 3 \newline

   LLM A provides more details about the property, such as mentioning the presence of mountain views, a babbling creek, pasture land, and a storage shed. On the other hand, LLM B's response lacks specific details about the land itself, instead focusing more on the nearby town and activities. \newline

   . \newline
   . \newline
   . \newline

7. Accuracy of information: \newline
   - LLM A: 5 \newline
   - LLM B: 5 \newline

   Both responses provide information that does not require fact-checking. They introduce no factual inconsistencies. \newline

Based on the evaluation above, my verdict is: \newline

[[B]]: 3, 3, 5, 5, 4, 3, 5

\end{tabular}
\end{center}
\end{table}

\begin{table}[h]
\caption{Simple judger prompt and completion from GPT-3.5-Turbo for the example response set given in Table \ref{example_responses}. In this example, the randomization resulted in LLM A being the unwatermarked completion, and LLM B being the watermarked completion.}
\label{simple_judger_example}
\begin{center}
\begin{tabular}{p{0.5\textwidth}p{0.5\textwidth}}
\multicolumn{1}{c}{\bf SIMPLE PROMPT}  & \multicolumn{1}{c}{\bf JUDGER RESPONSE}
\\ \hline \\
Please act as an impartial judge and evaluate the quality of the text completions provided
    by two large language models to the prompt displayed below. You should choose the LLM that
    completes the prompt better. Begin your evaluation by comparing the two responses and provide
    a short explanation. Avoid any position biases and ensure that the order in which the responses
    were presented does not influence your decision. Do not allow the length of the responses to
    influence your evaluation. Be as objective as possible. After providing your explanation, output
    your final verdict by strictly following this format: "[[A]]" if assistant A is better, "[[B]]"
    if assistant B is better, and "[[C]]" for a tie.

[Prompt] \newline

[LLM A’s Answer] \newline

[LLM B’s Answer] \newline

& The response provided by LLM A provides more specific information about the property and its potential uses. It mentions that the land could be used as a homestead, farm, or a place to park an RV. It also highlights the proximity to the Blue Ridge Parkway, Boone, and Watauga River, appealing to outdoor enthusiasts. Additionally, it mentions that the property is being sold as is with no disclosures or warranties, and that a cash deal is the most likely outcome.\newline

On the other hand, LLM B's response is focused more on the location and nearby amenities. It mentions that the property is located in a quiet and peaceful area, with nearby access to the Blue Ridge Parkway and various outdoor activities. It also briefly mentions the nearby town of Spruce Pine and the options for shopping, dining, and entertainment. The response concludes with directions to the property from a specific intersection.\newline

Overall, both responses provide relevant information, but LLM A's response provides more details about the property and its potential uses. Therefore, based on the evaluation, [[A]] is better.

\end{tabular}
\end{center}
\end{table}

\newpage 
~\newpage

\section{Appendix 2: Binary Classifier}
\label{sec:classifier_details}

The classifier is a multi-layer perceptron (MLP) with the following architecture:
\begin{itemize}
    \item Input size: 1536
    \item Output size: 1
    \item Hidden layers: 512, 256, 128, 64
    \item Activation: ReLU between layers, sigmoid at the output
\end{itemize}
Optimization was performed using the Adam optimizer with: \(\beta_1 = 0.5\), \(\beta_2 = 0.999\), learning rate and weight decay determined through a hyperparameter grid search (parameters shown in Table \ref{hyperparameter_search}). A dynamic learning rate scheduler was used (\texttt{torch.optim.lr\_scheduler.ReduceLROnPlateau}) with a factor of 0.5 and a patience of 50 epochs. Training was conducted for 150 epochs. \\

\begin{table}[h]
\centering
\begin{tabular}{ll}
\hline
\textbf{Hyperparameter} & \textbf{Value} \\
\hline
Adam Weight Decay & Grid search over $\{2e^{-4}, 2e^{-3}, 2e^{-2}\}$\\
Learning Rate & Grid search over $\{2e^{-5}, 2e^{-4}, 2e^{-3}\}$ \\
Batch Size & Grid search over $\{50, 75, 100\}$ \\
Dataset Randomization & Grid search over \{On, Off\}  \\
\hline
\end{tabular}
\label{hyperparameter_search}
\caption{Binary classifier hyperparameters}
\end{table}

\end{document}